\documentclass[10pt,twocolumn,letterpaper]{article}

\usepackage{iccv}
\usepackage{times}
\usepackage{epsfig}
\usepackage{graphicx}
\usepackage{amsmath}
\usepackage{amssymb}
\usepackage{verbatim}
\usepackage{multirow}
\usepackage{subfig}
\usepackage{theorem}
{\theorembodyfont{\rmfamily} \newtheorem{Remark}{Remark}}

\usepackage[pagebackref=true,breaklinks=true,letterpaper=true,colorlinks,bookmarks=false]{hyperref}

\iccvfinalcopy 

\def\iccvPaperID{9} 

\ificcvfinal\fi
\begin{document}

\title{NO Need to Worry about Adversarial Examples in Object Detection in Autonomous Vehicles}

\author{Jiajun Lu\thanks{These authors contributed equally}, Hussein Sibai\footnotemark[1], Evan Fabry, David Forsyth\\
University of Illinois at Urbana Champaign\\
\{jlu23, sibai2, efabry2, daf\}@illinois.edu
}

\maketitle

\begin{abstract}
	It has been shown that most machine learning algorithms are susceptible to adversarial perturbations. Slightly perturbing an image in a carefully chosen direction in the image space may cause a trained neural network model to misclassify it. Recently, it was shown that physical adversarial examples exist:  printing perturbed images then taking pictures of them would still result in misclassification. This raises security and safety concerns. 

However, these experiments ignore a crucial property of physical objects:  the camera can view objects from different distances and at different angles.  In this paper, we show experiments that suggest that current constructions of physical adversarial examples do not disrupt object detection from a moving platform.  Instead,  a trained neural network classifies most of the pictures taken from different distances and angles of a perturbed image  correctly. 
We believe this is because the adversarial property of the perturbation is sensitive to the scale at which the perturbed picture is viewed, so (for example) an autonomous car will misclassify a stop sign only from a small range of distances. 

Our work raises an important question: can one construct examples that are adversarial for many or most viewing conditions? If so, the construction should offer very significant insights into the internal representation of patterns by deep networks.  If not, there is a good prospect that adversarial examples can be reduced to a curiosity with little practical impact.
\end{abstract}

\section{Introduction}
Deep neural networks (NN) are being successful in accomplishing challenging tasks such as image classification \cite{Krizhevsky2012imagenet} and speech recognition \cite{Hinton2012SpeechRecognition}. Given the architecture of the neural net, its parameters are chosen by a training algorithm based on a finite set of the input space, called the training set, and is expected to generalize. 

However, a carefully chosen small perturbation of the input may cause a network to result in a different answer \cite{szegedy2013intriguing, fawzi2015fundamental, nguyenfooled}. In that case, the new input is called an adversarial example. For instance, one can perturb an image to cause a NN to misclassify it while keeping the change small enough to not be perceptible to a human eye. Even worse, these perturbations were found to generalize over different NN architectures and training datasets. This means that an attacker can train a classifier and use it to generate adversarial version of an image, then use it to fool another model.

 In the past few years, researchers tried to explain why neural nets are susceptible to such examples despite their impressive success on random test datasets \cite{goodfellow2014explaining, fawzi2016robustness, fawzi2015analysis}, suggested new methods to generate adversarial examples and measure NN robustness against them \cite{DBLP:journals/corr/FawziMF16, DBLP:journals/corr/SabourCFF15,DBLP:journals/corr/Moosavi-Dezfooli15, DBLP:journals/corr/KatzBDJK17, moosavi2016universal, papernottransfer, Song2016}, and proposed ways to improve the networks' robustness against these examples \cite{DBLP:journals/corr/GuR14, metzen2017detecting, lu2017safetynet}. In all of these cases, the adversarial perturbation was added to a digital image then fed as input to the neural network. 
 
 Then, the natural question is if these perturbed images do stay adversarial if taken as input from the physical world using a camera. They actually do as shown in \cite{DBLP:journals/corr/KurakinGB16}. There, the authors generated adversarial examples of images from the ImageNet dataset using the fast method presented in \cite{goodfellow2014explaining} and two iterative versions of it. They then printed these examples and took pictures of them using cellphone cameras. The images were passed as input to trained neural networks to demonstrate that the output is still misclassified. That shows that the adversarial perturbations were robust to the transformations and the noise resulting from the camera and phone processing.
 
 Another physical attack against face detection systems was presented in \cite{Sharif:2016:ACR:2976749.2978392}. The authors demonstrated both black and white box methods to print sun glasses that cause a state-of-the-art face recognition system to misclassify the attacker face to a specific or arbitrary other face. Similar to \cite{DBLP:journals/corr/KurakinGB16}, the photos were taken from a short distance and a single photo was taken for each image.

These experiments raise serious safety and security concerns, especially when these networks are incorporated in safety-critical systems such as autonomous vehicles. For example, adding subtle adversarial perturbations, which humans generally don't notice, to a stop sign to cause it to be misclassified as a minimum speed limit sign could lead to fatal car accidents. 

In this paper, we present experiments that show physical adversarial examples do not generalize over several distances and angles for object detector. Specifically, we show that no matter how easy or hard to perceive the physical adversarial perturbations added to stop signs are, most of the video frames taken from a camera in a car approaching the adversarial stop signs will be correctly classified. As a result, the controller of an autonomous car taking the frames as input will make the right decision most of the time.

\section{Methods for Generating Adversarial Examples}
\label{Sec:methods}
We used the following weak and strong attacks~\cite{carlini2016defensive}, with various choices of
hyper-parameters, to test the robustness of two neural networks with different architectures. Given a trained NN, each attack searches for an image $\pmb{X}_{adv}$ similar, from human perspective, to the target image $\pmb{X}$ which 
changes the NN classification of $\pmb{X}$ from class $y_{true}$ (represented as one hot vector) to either an arbitrary or specifically chosen class $y_{fool}$. 
 An image $\pmb{X}$, is a 3-dimensional tensor representing width, length, and depth. The depth is assumed to have an integer value between 0 and 255. We follow the literature in calling the images resulting from running these methods on some input images {\em adversarial}, even if it is not guaranteed that they will not be classified as $y_{true}$. Finally, we denote by $J(\pmb{X}, y)$ the cross-entropy cost function where the input image is $\pmb{X}$ and its true class is $y$. $J$ is used to train both of the neural networks that we will be considering. Also, $J$ is used along with its gradient in the different methods below to generate adversarial examples.

\subsection{Fast Sign method}
Goodfellow et al~\cite{goodfellow2014explaining} described this simple method of generating adversarial examples. 
The applied perturbation is the direction in image space which yields the
highest increase of the linearized cost under $l_{\infty}$ norm. It uses a hyper-parameter $\epsilon$ to govern the 
distance between adversarial and original image. Specifically, 
\[
\pmb{X}_{adv} = \pmb{X} + \epsilon \cdot \mbox{sign} (\nabla_{\pmb{X}} J(\pmb{X}, y_{true}) ).
\] 
when the target fooling class is known as $y_{fool}$, another update formula can be used.
\[
\pmb{X}_{adv} = \pmb{X} - \epsilon \cdot \mbox{sign} (\nabla_{\pmb{X}} J(\pmb{X}, y_{fool}) ).
\] 
In our experiments, we used $\epsilon = 0.2 \times 255 = 51$. This is very large value for $\epsilon$ with respect to the $0.1$ and 0.25 values which have been used in \cite{goodfellow2014explaining}. Higher $\epsilon$ values result in larger perturbation of the image. We want to show that even under the assumption of a strong attack, the adversarial properties would not hold across distances and angles, which means the photo will be correctly classified in the context of an autonomous vehicle.

\subsection{Iterative methods} 
Kurakin et al.~\cite{DBLP:journals/corr/KurakinGB16} introduced an iterative version of the fast sign method by applying it several times with a smaller step size, $\alpha$, and clipping all pixels after each iteration
to ensure that results stay in the $\epsilon$ neighborhood of the original image. 
 Formally, 

\begin{align*}
&\pmb{X}_{adv,0} = \pmb{X},\\
&\pmb{X}_{adv,N+1} = Clip_{\pmb{X},\epsilon} \{ \pmb{X}_{adv,N} \\
&\hspace{1.3in}+ \alpha \cdot \mbox{sign} (\nabla_{\pmb{X}} J(\pmb{X}_N, y_{true}) ) \}.
\end{align*}

where 
\begin{align*}
Clip_{\pmb{X},\epsilon}\{\pmb{X}'\} &= \min\{\ 255, \pmb{X}(x,y,z) + \epsilon, \\
&\hspace{0.3in} \max\{0, \pmb{X}(x,y,z) - \epsilon, \pmb{X}'(x,y,z)\} \}
\end{align*}
and $\pmb{X}(x,y,z)$ is the value of channel $z$ of $\pmb{X}$ at pixel of coordinate $(x,y)$. Hence, $Clip_{\pmb{X},\epsilon}\{\pmb{X}'\}$ keeps the values in $\pmb{X}'$ so that it stays in the $L_{\infty}$ $\epsilon$-ball around $\pmb{X}$. In our experiments, we set $\alpha$ to 10 and the number of iterations to $20$. These choices are heuristic.  

\subsection{L-BFGS method}
Szegedy et al \cite{szegedy2013intriguing} suggested that in order to compute an adversarial example, one could perform a line search using the box-constrained L-BFGS method to find the perturbation tensor $\pmb{R}$ that minimizes:
\[
c |\pmb{R}| + J(\pmb{X} + \pmb{R}, y_{fool}),
\]
where $c$ is a weight parameter, $\pmb{R}$ is the perturbation and $\pmb{X} + \pmb{R}$ is classified with some class $y_{fool}$ different than $y_{true}$. It is similar to the iterative method, except that the step size is estimated by L-BFGS, and no pixel clipping is applied. 

\subsection{Attacking a detector}
\label{sec:atd}
The current literature primarily investigates attacks against classifiers, i.e. generating adversarial examples that fool classifiers. In this paper, we extend it to attacking the YOLO multiple object detector~\cite{redmon2016you}~\footnote{\url{https://github.com/thtrieu/darkflow}}. When we feed an image as an input to the neural network part of the detector, the YOLO network outputs the vector $p_{output}$. In the case of a classifier, we use the  cross-entropy cost function between the output ($p_{output}$) and the one hot vector $y_{fool}$ (or $y_{true}$). In the case of a detector, there is no such one hot vector, $y_{fool}$, and we set the vector $y_{fool}$ to have same length as $p_{output}$ (write $l_{p_{output}}$), and each element has the same value of  $1/l_{p_{output}}$. Because $l_{p_{output}}$ is relatively large, each element of $p_{output}$ approaches zero, meaning that nothing is detected. In summary, to attack the multiple object detector we replace $y_{fool}$ with a near zero vector and otherwise follow the recipe for attacking a classifier. This adversarial attack against the detector may seem a bit odd, but it can generate adversarial images that fool detectors reliably.  We used this method to attack detectors for all results in this paper. 

Generally, the L-BFGS method makes smaller modifications to the input but has a higher rate of successful attacks.
Similarly, the Iterative method outperforms the Fast Sign method.  This is for attacks that consist of image modifications; in what
follows, we focus on physical attacks (i.e. those made by imaging real objects).

\begin{figure*}
	\centerline{\includegraphics[width=0.7\linewidth]{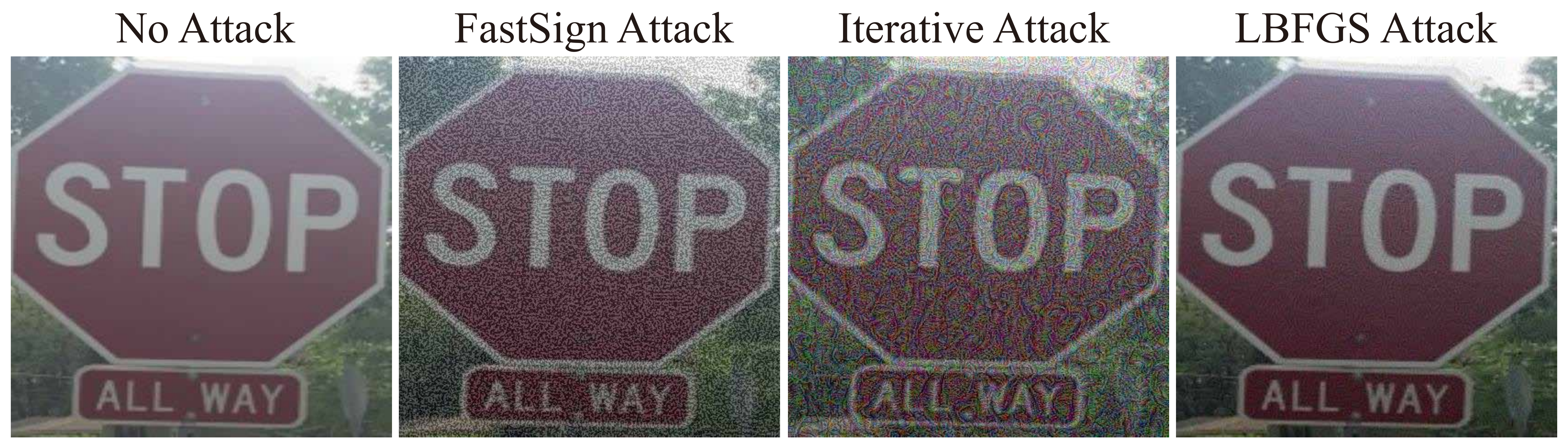}}
	\caption{Adversarial examples for YOLO detector. In this case, adversarial perturbations have relatively high frequency, so the perturbations are more visible from near distance and cannot be preserved for far distance. }
	\label{fig:detector_adv}
\vspace{-1em}
\end{figure*}

\begin{figure*}
	\centerline{\includegraphics[width=0.7\linewidth]{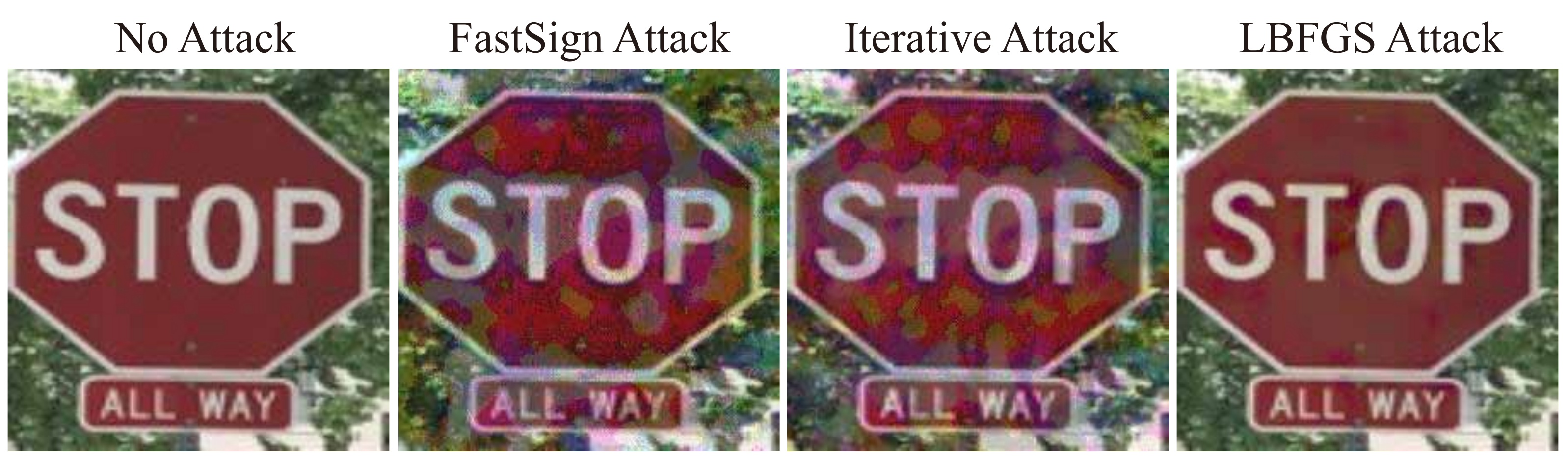}}
	\caption{Adversarial examples for traffic sign classifier.  In this case, adversarial perturbations have relatively low frequency, so the perturbations are relatively more visible from far distance.  }
	\label{fig:classifier_adv}
\vspace{-1em}
\end{figure*}

\section{Physical Adversarial Examples}
\label{sec:physical}

Physical adversarial examples differ from virtual adversarial examples because the camera position is now not controlled.  This means, for example, the 
distance from the camera to the physical object can change, changing the pattern of pixels in the resulting image.  Our experiments suggest that this disrupts
most attacks severely.  We focus on distance between camera and object, but show some results from camera rotation below as well.

We follow \cite{DBLP:journals/corr/KurakinGB16} in using the destruction rate to evaluate the effectiveness of the methods to generate adversarial examples in the physical world. Destruction rate measures the percentages of perturbed images that are no longer adversarial in the physical world. It was defined as follows:

\[
d = \frac{\sum_{k=1}^{n} C(\pmb{X}^k, y_{true}^k)\overline{C(\pmb{X}_{adv}^{k},y_{true}^k)}C(T(\pmb{X}_{adv}^k), y_{true}^k)}{\sum_{k=1}^{n} C(\pmb{X}^k, y_{true}^k)\overline{C(\pmb{X}_{adv}^k, y_{true}^k)}},
\]
where $n$ is the total number of images, $\pmb{X}^k$ is the $k^{th}$ image, $y_{true}^k$ is its true class, and $\pmb{X}_{adv}^k$ is the perturbed version of $\pmb{X}^k$ generated by the method that we are evaluating, and $T(\cdot)$ represents an arbitrary transformation of the image represented here by printing it and taking a photo of it from some distance and angle. Finally, 
\[
	C(X,y) := 
	\begin{cases}
	1, \mbox{if image $X$ is classified as y}\\
	0, \mbox{otherwise},
	\end{cases}
\] 
and $\overline{C(X,y)} = 1 - C(X,y)$. 
If the destruction rate increases when the transformation changed from $T$ to $T'$, for example by changing the distance at which we take the photos, it means that the adversarial perturbation is less effective for $T'$.

\subsection{Controlled Experimental Setup}
\begin{figure*}
	\centerline{\includegraphics[width=0.8\linewidth]{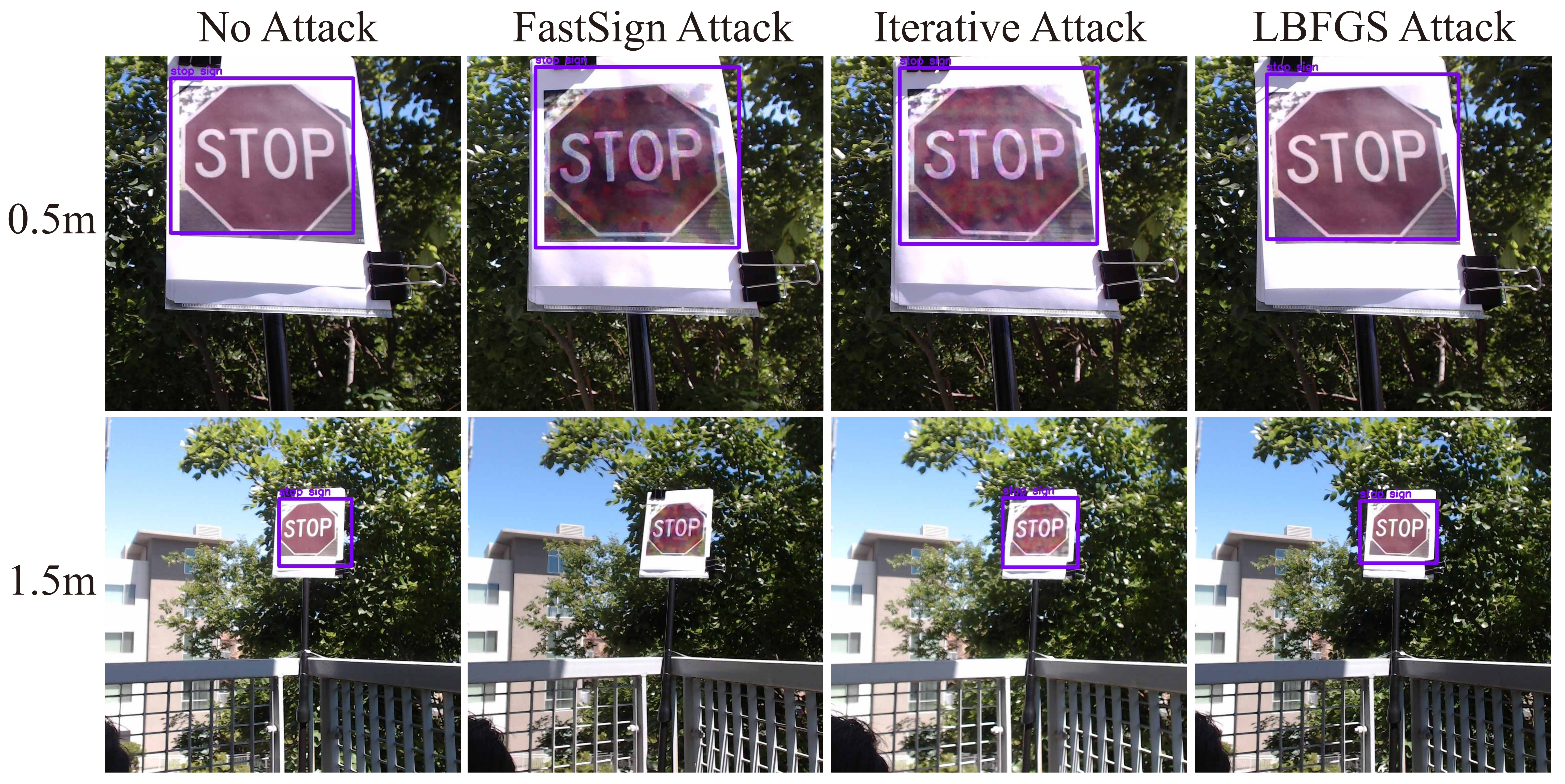}}
	\caption{This figure shows experiment setup, and we use the printed stop signs to simulate real stop signs with natural background. These are examples for successful 0.5 meters and 1.5 meters detection: both original images and adversarial examples are detected in both distances. It demonstrates that adversarial examples in a physical setting do not reliably fool stop sign detectors. }
	\label{fig:05m}
\vspace{-1em}
\end{figure*}

\begin{figure}
	\centerline{\includegraphics[width=0.9\linewidth]{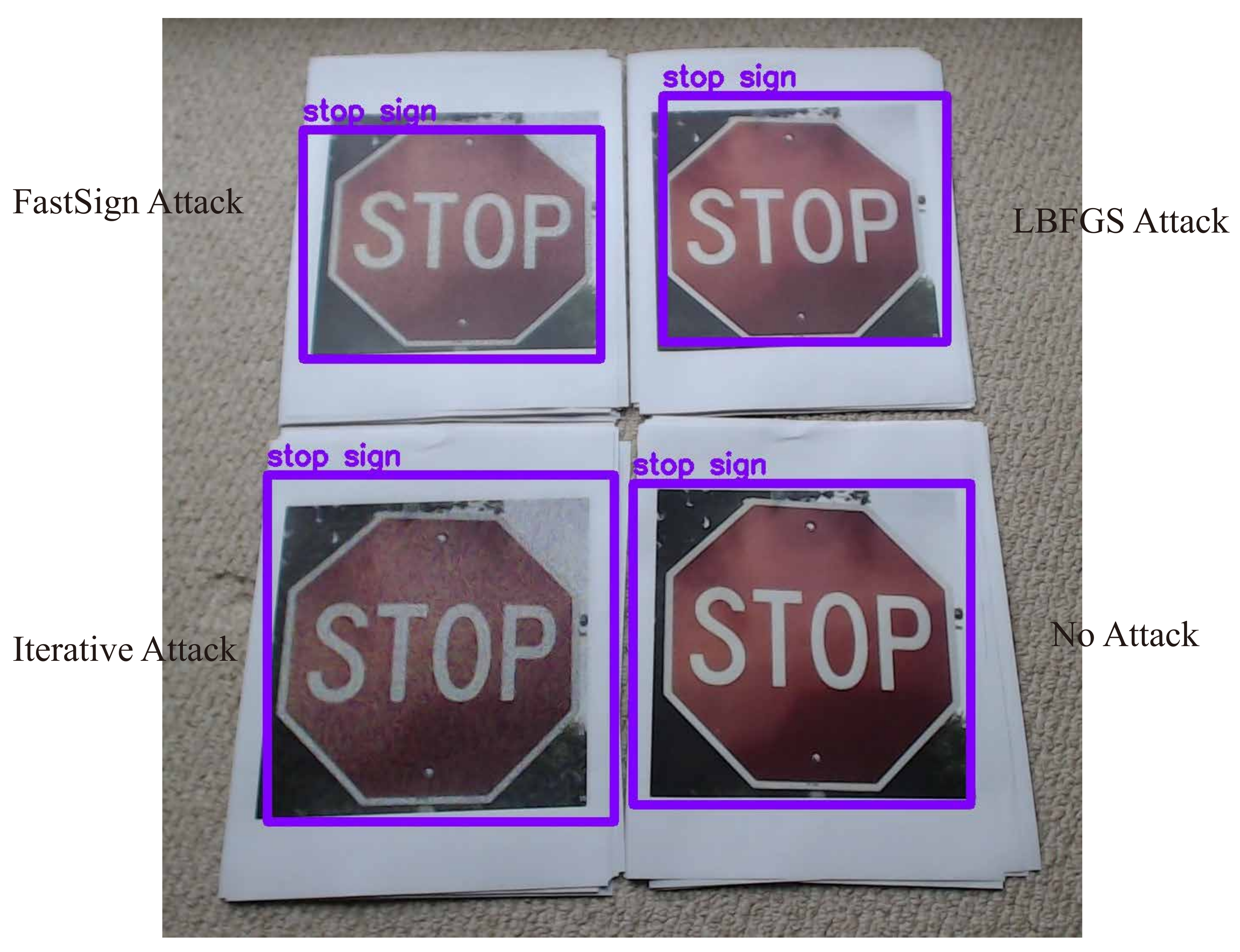}}
	\caption{Example for detecting four images together. One original non adversarial stop sign and three different adversarial stop signs are all detected as stop signs. It means adversarial attacks on sign detectors fail. }
	\label{fig:grid}
\vspace{-1em}
\end{figure}

We generated adversarial examples against both the YOLO detector and the traffic sign classifier, and then printed out these adversarial images. 
We simulated the process of a car driving by the printed signs, and checked the YOLO detector's detection rate. Specifically, we did the following steps:
\begin{enumerate}
\item We drove and took 180 photos of stop signs with an iPhone 7 from the passenger side window of the car. The photos were taken from a variety of different angles and distances under different lighting conditions. These samples represent the average case in \cite{DBLP:journals/corr/KurakinGB16}. They are random samples generated by normal driving conditions. They were not meant to represent extreme weather and lighting conditions.

\item We installed the pretrained open-source object detector Darkflow (YOLO). The model is pretrained on MS-COCO dataset~\cite{lin2014microsoft}, which includes the class of stop sign. For each of the first 100 images collected, we generated three adversarial images based on the darkflow detector using the three different methods described in Section~\ref{Sec:methods}.

\item We trained a VGG16 traffic sign classifier from scratch using the German traffic signs dataset\footnote{\url{http://benchmark.ini.rub.de/}}. However, that classifier was not able to recognize US stop signs. To solve that issue, we added the first 150 of our collected cropped stop sign images to the training set. For the remaining 30 images, we generated adversarial examples using the three different methods in Section~\ref{Sec:methods}s.

\item We computed the detection rate on the full size non-cropped digital original images and perturbed images. The results are shown in the third column in table \ref{table:results}.

\item We manually cropped the stop signs from the first 100 collected clean images and generated three perturbed versions for each one of them against the detector. An example of a cropped stop sign and its perturbed versions are shown in Figure \ref{fig:detector_adv}. We checked the detection rate on these 400 digital images and the results are shown in the first four rows of the fourth column in table \ref{table:results}.  We also cropped and generated adversarial examples for the last 30 images against the classifier, and the results are shown in the last four rows in the same column. An example of a cropped clean sign and its perturbed versions against the classifier are shown in Figure \ref{fig:classifier_adv}.

\item We printed the resulting $(100 + 30) \times 4$ cropped stop signs' images on A4 papers, then took photos of them from two distances: 0.5m and 1.5m (Figure~\ref{fig:05m}). Note that we chose these distances since A4 papers (8.27 in $\times$ 11.7 in) are much smaller the actual US stop signs (18 in $\times$ 18 in, 24 in $\times$ 24 in, or 30 in $\times$ 30 in) which means that taking a photo from these distances is equivalent to taking the photo of a real stop sign from few meters as a car camera would do. Then, we computed the detection rate on these retake photos. The results are shown in the sixth column in table \ref{table:results}. We show another example of failed physical adversarial attack with four printed signs together in Figure~\ref{fig:grid}. (The photos were taken using a Logitech C922 Pro camera; the printer used is imageRUNNER ADVANCE C5030.)

\end{enumerate}

\begin{Remark}
Our results suggest that physical attacks on a detector are ineffective, likely because of the camera movements (changes in ``distance''). To control for the possibility that
our attack on the detector might have unanticipated consequences, we also attacked a classifier.  Note that, in this case, distance makes attacks ineffective as well.
\end{Remark}

\subsection{Controlled Experiment Results and Analysis}
There are four metrics of classification accuracy that we care about: accuracy on clean images, on perturbed ones, and on the photos taken of the printed clean and perturbed images. Refer to Table~\ref{table:results}.

\begin{table*}[h!]
	\begin{center}
		\resizebox{1.0\textwidth}{!}
		{
			\begin{tabular}{  c || c || c | c  || c || c | c | c }
				Adversarial Target  & Images & Ori DR\% & Crop DR\% & Distance &  Physical DR\% & Relative DR\% & Destruction Rate\%\\
				\hline
				\multirow{8}{*}{Detector} & \multirow{2}{*}{Original} & \multirow{2}{*}{30\%} & \multirow{2}{*}{11\%} & 0.5m & 65\% & 100\% & NA\\
				&  &  &  & 1.5m & 28\% & 100\% & NA\\
				\cline{2-8}
				&  \multirow{2}{*}{ Fast Sign Adv} & \multirow{2}{*}{8\%} & \multirow{2}{*}{0\%} & 0.5m & 57\% & 88\% & 78\%\\
				& &  &  & 1.5m & 29\% & 104\% & 82\%\\
				\cline{2-8}
				& \multirow{2}{*}{ Iterative Adv} &  \multirow{2}{*}{0\%} & \multirow{2}{*}{0\%} & 0.5m & 29\% & 45\% & 32\%\\
				&  &  &  & 1.5m & 27\% & 96\% & 86\%\\
				\cline{2-8}
				& \multirow{2}{*}{ LBFGS Adv} &  \multirow{2}{*}{6\%} & \multirow{2}{*}{3\%} & 0.5m & 34\% & 52\% & 51\%\\
				&  &  &  & 1.5m & 33\% & 118\% & 93\%\\
				\hline
				\hline
				\multirow{8}{*}{Classifier} & \multirow{2}{*}{Original} & \multirow{2}{*}{37\%} & \multirow{2}{*}{14\%} & 0.5m & 67\% & 100\% & NA\\
				&  &  &  & 1.5m & 30\% & 100\% & NA\\
				\cline{2-8}
				&  \multirow{2}{*}{ Fast Sign Adv} & \multirow{2}{*}{NA} & \multirow{2}{*}{0\%} & 0.5m & 47\% & 70\% & 70\%\\
				&  &  &  & 1.5m & 7\%  & 23\% & 22\%\\
				\cline{2-8}
				& \multirow{2}{*}{ Iterative Adv}  &  \multirow{2}{*}{NA} & \multirow{2}{*}{0\%} & 0.5m & 40\% & 60\% & 60\%\\
				&  &  &  & 1.5m & 20\% & 67\% & 67\%\\
				\cline{2-8}
				& \multirow{2}{*}{ LBFGS Adv} &  \multirow{2}{*}{NA} & \multirow{2}{*}{3\%} & 0.5m & 67\% & 100\% & 90\%\\
				& &  &  & 1.5m & 34\% & 113\% & 100\%\\
				\hline
			\end{tabular}
		}
		\vspace{0.5ex}
		\caption{This table summarizes the physical adversarial attack experiments. All the evaluations are performed with pretrained YOLO detector, and DR means stop sign detection rate. The top half of the table summarizes the adversarial examples generated from the detector, and the bottom half of the table summarizes the adversarial examples generated from the classifier. Ori DR denotes the detection rate on the full scene images with stop signs. Crop DR denotes the detection rate on images that are cropped close to the edges of the stop signs. We print the original and various adversarial stop signs out, and put them in different distances. Physical DR means the detection rate on these retaken images (see Figure~\ref{fig:05m}). Relative DR means the relative detection rate between the same setting adversarial images and original images. Destruction Rate is calculated as Section~\ref{sec:physical}. From the table, we can tell that the destruction rate of adversarial examples for detector in 1.5 meters is high, and that in 0.5 meters might be high. It means the physical detector adversarial attacks are not successful across different distances, especially when the distance is far and the small patterns are not visible. The destruction rate of adversarial examples for the classifier varies by distance for the Fast Sign attack, remains relatively high for the Iterative attack, and is high for the LBFGS attack. This suggests that physical classifier adversarial attacks are also not successful across different distances. }
		\label{table:results}
		\vspace{-2ex}
	\end{center}
\end{table*}

\begin{figure}
	\centerline{\includegraphics[height=2.5in,width=0.6\linewidth]{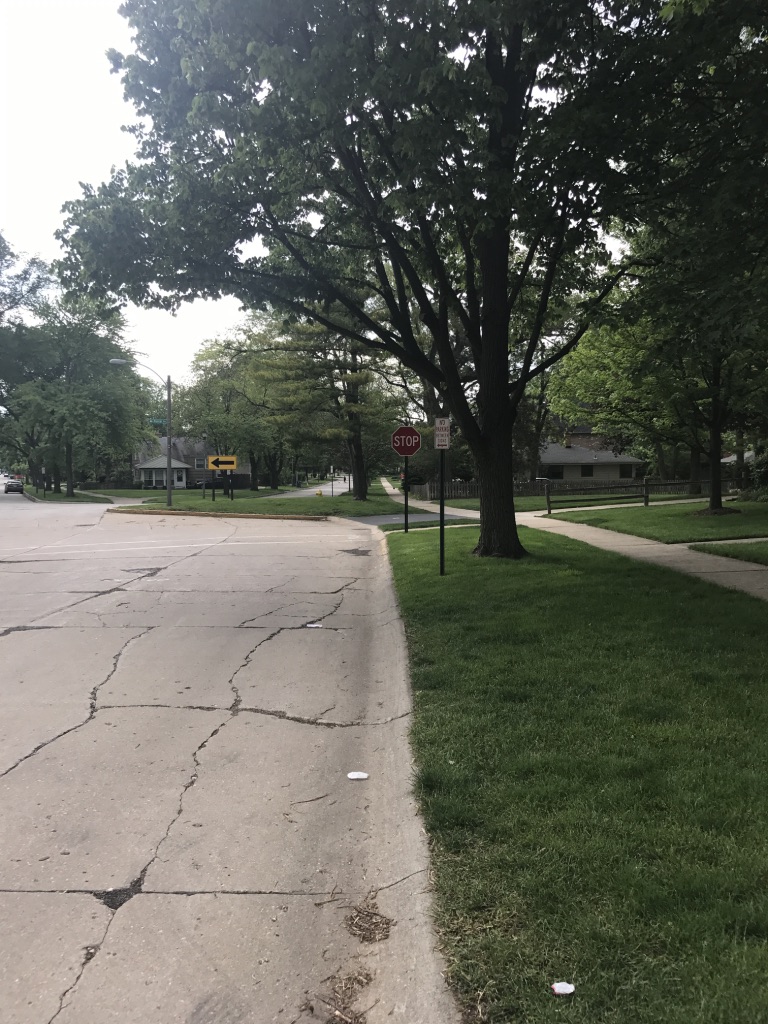}}
	\caption{A far stop sign. The image will be further resized and fed into a YOLO detector, while YOLO is not good at detecting small objects. As a result, detecting the stop sign would be hard. }
	\label{fig:farcleanstopsign}
\vspace{-1em}
\end{figure}

The detection rate on the clean photos was only $30 \%$ and that is because in many of the images, the stop signs were far and not easily detectable by the detector. An example of these images is shown in Figure \ref{fig:farcleanstopsign}. Furthermore, adding adversarial perturbations to the images decreased the detection rate significantly (see column 3 in table \ref{table:results}). 

One may expect that cropping the stop signs from the images would help the detector recognize the stop sign. However, the detection rate decreased significantly after cropping as shown in the fourth column in table \ref{table:results}. The clean images, which were misclassified when they were full images, are probably images of signs taken from a distance or under poor lighting conditions. Cropping these images and rescaling them did not add extra features or improve the conditions for the detector, so it is reasonable that they remain misclassified. Cropping the stop signs from correctly classified images removed the background, resulted in some of them being misclassified. This suggests that the background provides some feature information to the detector. The decrease in the detection rate for the perturbed versions of the cropped signs is expected. In these cases, in addition to the background issue, the adversarial perturbation is focused on the stop sign rather than the whole non-cropped image. So with the same adversarial target, adding adversarial perturbation after cropping results in a stronger attack than adding it before.

Surprisingly, printing these cropped images and taking photos of them increased the detection rate significantly. For example, the detection rate on the photos taken of clean images from a 0.5m distance is 65\%, which is much larger than that on the clean cropped images (11\%). This suggests that the performance of the darkflow detector at identifying stop signs improves when there is a high contrast background rather than when the object dominates the frame.

Moreover, as expected, taking the photos from further away decreased the detection rate of the detector. This can be seen in the last column of table \ref{table:results}. As an example, the detection rate on the photos of the clean images decreased from 65\% to 28\% when we took the photos from a distance of 1.5m instead of 0.5m. 

The major idea of the paper is that distance will corrupt the adversarial perturbation. This can be seen in the last two columns of table \ref{table:results}. For instance, the relative detection rate between the photos taken of the images resulting from the Iterative method and the original images increased from $29/65 = 45\%$ at 0.5m to $27/28 = 96\%$ at 1.5m. Similarly the destruction rate, increased from 32\% to 86\%. The increase when we consider the Fast Sign method instead of the Iterative method is smaller (from 88\% to 104\%) since it is already a weak attack; the detection rate is 57\% for photos taken from 0.5m of images generated using the Fast Sign method. 

Notice that it is not necessary that the relative detection rate increases as distance increases, it is only the case that the adversarial perturbation is most effective at some particular distance and its effectiveness decreases for other distances. For example, the relative DR decreased from 70\% to 23\% when the photos of images produced by the Fast Sign method, attacking the classifier, were taken from a 1.5m distance instead of 0.5m. In that case, the adversarial perturbation was more effective at 1.5m than at 0.5m. This is likely because the adversarial perturbations generated from attacking the classifier were generally coarser and thus more recognizable from a distance (see Fig. \ref{fig:detector_adv} and Fig. \ref{fig:classifier_adv}).

Finally, the key idea is that if an adversary added a physical perturbation to a stop sign, since the camera has a fixed resolution, the sign will be misdetected by an autonomous car detector at a certain distance, but it will still be detected from other distances. Thus, it will not be a dangerous threat as the control system should identify that there is a stop sign by a simple rolling majority voting algorithm over the frames. As attacks are weaker at distance, the rolling majority voting algorithm should begin with many frames in which the stop sign is correctly identified once the vehicle passes the maximum detection distance for the detector. The details of an experiment to check what happens in real life in such situations is shown in the next section.

\begin{figure*}
	\centerline{\includegraphics[width=0.8\linewidth]{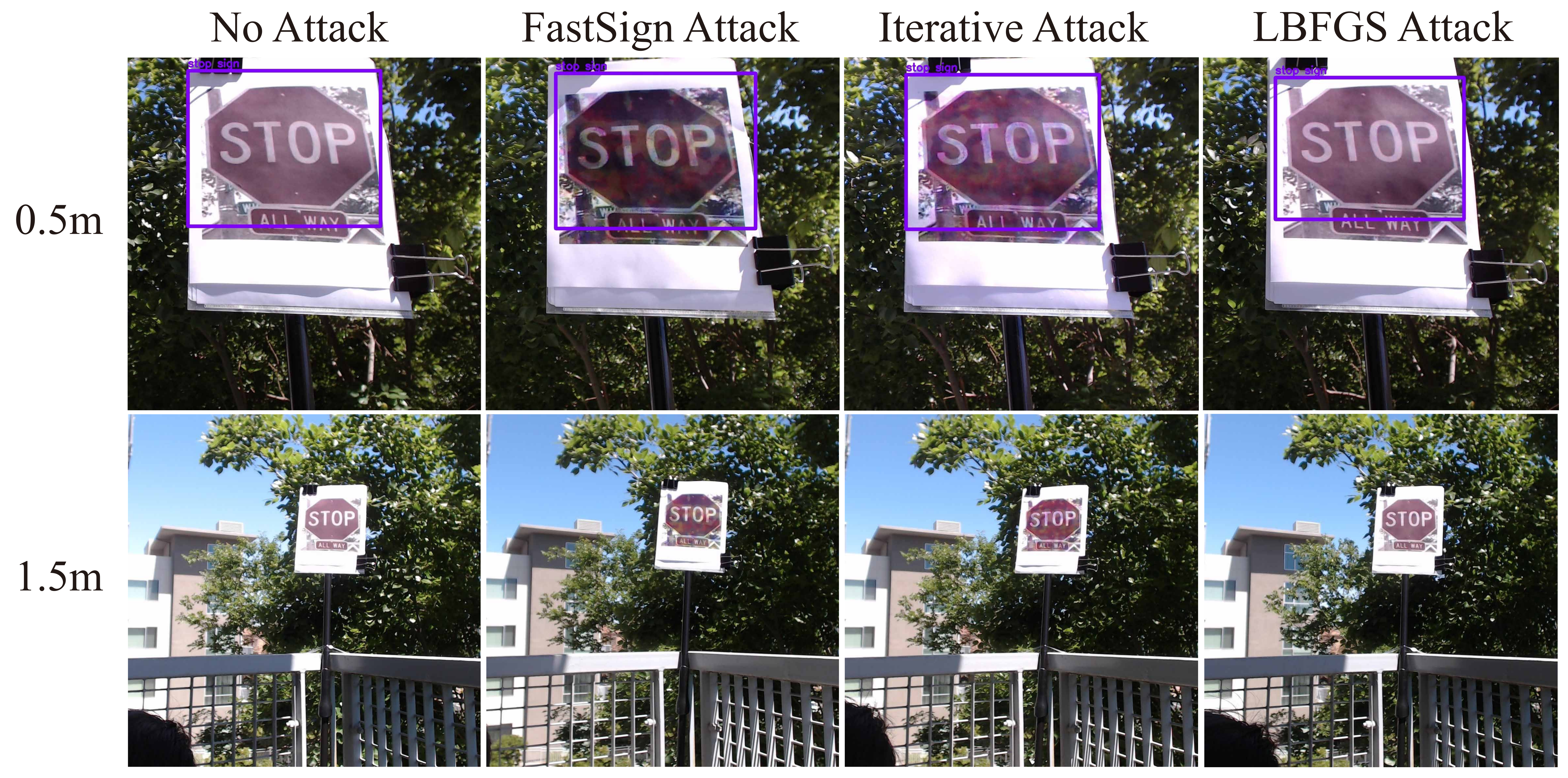}}
	\caption{Example of the effects of distance. Stop signs are detected for non-adversarial examples and all adversarial examples when the distance is 0.5 meters, however none are detected at a distance of 1.5 meters. Small stop signs in the image are difficult for YOLO to detect regardless of whether they are adversarial examples or not. }
	\label{fig:distance}
\vspace{-1em}
\end{figure*}

\section{Experiment in Real Life}
In this section, we describe an experiment we performed to test the ability of different adversarial methods in Section \ref{Sec:methods} to cause an autonomous car to misdetect a stop sign.

The experiment consists of the following steps:
\begin{enumerate}
\item We printed eight images corresponding to two stop signs along with their adversarial versions attacking the detector on two $34$in $\times 44$in posters using an HP DesignJet Z5600 printer and cut them to individual signs. We also printed one image with its adversarial versions on two posters of the same size and using the same printer.
\item We attached these signs to an actual stop sign one at a time.
\item We attached a Logitech C922 Pro camera to the wind shield of a Lincoln MK-Z car and connected it to a server in the car with darkflow installed.
\item We started recording video with 2 frames/second rate while running the detector in real time.
\item We drove toward the stop sign several times with different versions of printed stop signs attached. 
\end{enumerate}

There are several observations from the experiments:
\begin{itemize}
	
\item The printed stop signs (both original version and adversarial versions) are first detected from a closer distance than for an actual stop sign. As mentioned earlier, this is expected since the size of the printed sign is smaller and the image blurrier than actual stop signs, which are more reflective and have higher intensity.

\item Rarely were the printed perturbed stop signs misclassified, since as the car approaches the stop sign, the distance and angle change and eventually the stop sign disappears from the frames. In other words, the camera was not close enough to the stop sign to recognize the adversarial perturbation to make the sign misdetected.

\item In one case, a printed perturbed stop sign was misdetected as a sports ball in only two frames and it was detected correctly in the rest (see Figure~\ref{fig:misclassified_frisbee}). This misdetection most likely occurred due to the low resolution and natural detection error rather than due to our perturbation. In other cases, the printed adversarial sign and non-adversarial sign are similar. 
\begin{figure*}
	\centering
	\includegraphics[width=0.8\linewidth]{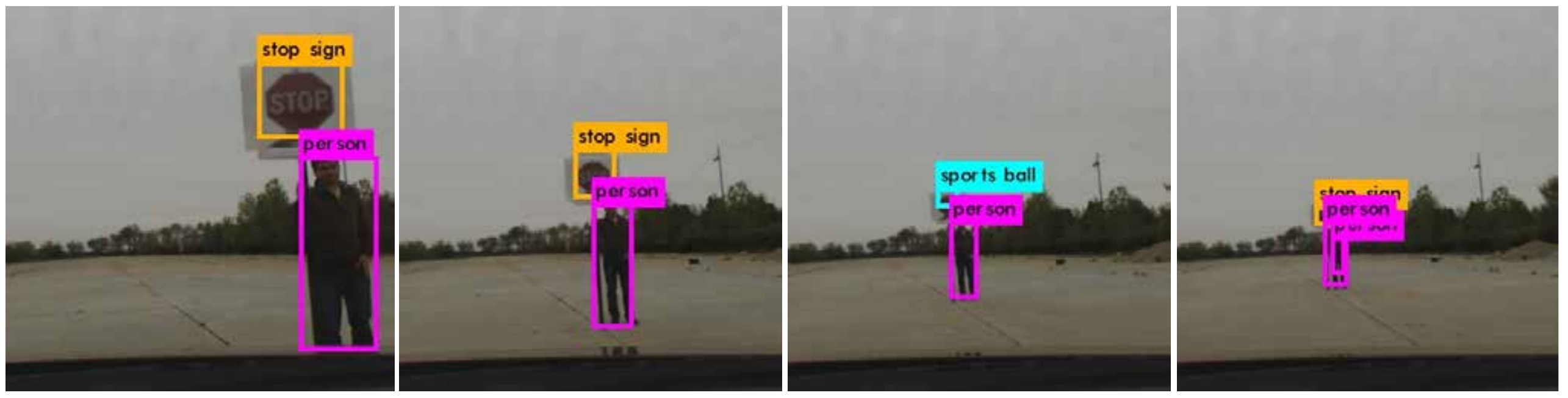}
	\caption{In the video sequence of autonomous vehicle approaching a printed perturbed stop sign, the stop sign is misclassified into a sports ball in two frames, and classified correctly in all other frames of various distances.}
	\label{fig:misclassified_frisbee}
\vspace{-1em}
\end{figure*}
\end{itemize}

\begin{figure*}
	\centering	
	\includegraphics[width=0.8\linewidth]{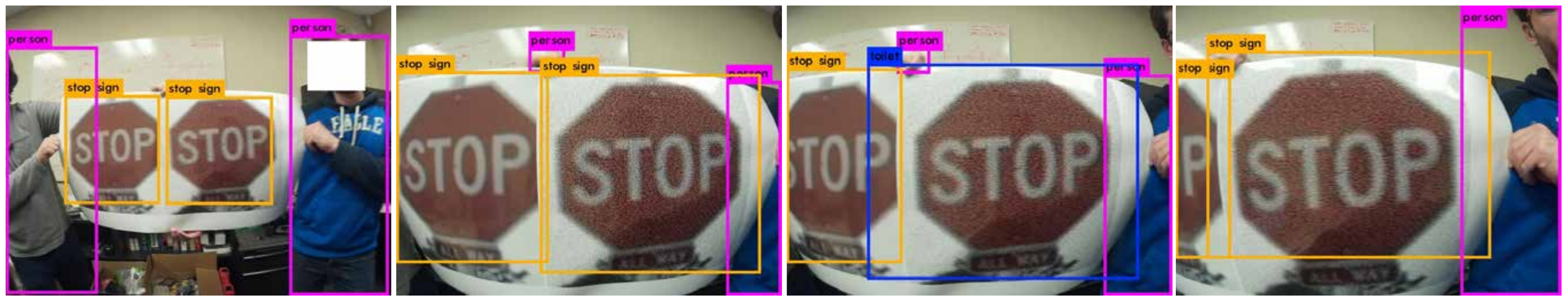}
	\caption{In the video sequence, the camera moves in close proximity around a printed original stop sign image and a printed adversarial stop sign. The original stop sign is always detected as stop sign while the adversarial stop sign is detected as toilet for two frames in an angle (detected as stop sign in all other angles). }
	\label{fig:misclassified_toilet}
\vspace{-1em}
\end{figure*}

In another experiment in the lab, we formed the hypothesis that even the angle from which we took the photo impact the effectiveness of the adversarial perturbation. A stop sign and a perturbed version using fast sign method were printed on a poster. Both were classified correctly from a distance between 20cm and 2m. When the distance between the camera and the poster is around 20cm, 
one capture angle consistently led darkflow to misdetect the perturbed image as a toilet while the clean sign remained correctly detected (see Figure~\ref{fig:misclassified_toilet}). This is important for our use case, since as a car approaches a stop sign, the angle of capture changes. Hence, if the perturbation generated by a method requires some angle of capture for it to be adversarial, it would most likely not form a significant threat for an autonomous car as it will be take images from different angles as it approaches the sign, so some of the frames will be detected correctly.

\section{Discussion}

For attacking detectors in Section~\ref{sec:atd}, YOLO regresses a flattened vector corresponding various estimates for each cell within a grid-partitioned image. Our adversarial generation method perturbs this entire vector which contains bounding box coordinates, object confidences and class confidences. Even though this attacking method can efficiently generate adversarial examples against the detector, it is not an ideal method of attack because it focuses a lot of effort on modifying unrelated objectives. In sum, this attack method might make our adversarial examples less relatively weak, so easier to detect in the real circumstances. In future works, we would like to identity the class confidence portion of the vector and generate attacks that only modify the stop sign class confidence values. 

Because the detector will make mistakes even without adversarial perturbations, it is important to distinguish between error resulting from the natural error rate of the detector and error due to adversarial effects. We have found it generally reliable to take multiple images in the same setting, as detector errors are quite random while adversarial effects are quite consistent. 

Our hypothesis that distance will affect adversarial examples is verified by many experiments, however we only performed a simple experiment to test our hypothesis that angles also impact the effectiveness of adversarial examples. In future works, we would like to design a comprehensive experiment to verify this angle hypothesis.

In our experiments, we used a drastically larger epsilon in all attacking methods. As known in the literature, larger epsilon could generate adversarial examples at higher rate and form possibly stronger attacks. We also performed experiments with small epsilon (not reported in the paper), and these adversarial attacks are even less effective in the physical world. The usage of larger epsilon makes it hard to compare against existing literature, but it makes the points of this paper clearer. 

Our work raises an important question: can one construct examples that are adversarial for many or most viewing conditions? If so, the construction should offer very significant insights into the internal representation of patterns by deep networks.  If not, there is a good prospect that adversarial examples can be reduced to a curiosity with little practical impact.

\section{Conclusion}

In this paper, we showed empirically that even if adversarial perturbations might cause a deep neural network detector to misdetect a stop sign image in a physical environment when the photo is taken from a particular range of distances and angles, they cannot reliably fool object detectors across a scale of different distances and angles. We collected images of stop signs and generated perturbed versions of them using three different adversarial attack methods, attacking both a classifier and an object detector. We used the YOLO detector to test and measure the detection rate on all of these images. Then, we printed them, took photos from different distances, and checked the fraction of the perturbed versions that did not remain adversarial (destruction rate) at each distance. The destruction rate is high in most cases and increases when the distance increases. Finally, we showed with a small experiment that the angle from which we take the photo can also change the effectiveness of an adversarial perturbation. In summary, existing adversarial perturbation methods applied to stop sign detection (with our dataset and controlled experiment) only work in carefully chosen situations, and our preliminary experiment shows that we might not need to worry about it in many real circumstances, specifically with autonomous vehicles.

{\small
\bibliographystyle{ieee}
\bibliography{egbib}
}

\end{document}